\documentclass{llncs}

\usepackage{graphicx}
\usepackage{amsmath}

\newcommand{\ie}{i.\,e.~}
\newcommand{\eg}{e.\,g.~}

\DeclareMathOperator{\argmin}{\text{arg\,min}}

\renewcommand{\vec}[1]{\mathbf{#1}}
\newcommand{\mat}[1]{\mathbf{#1}}

\newcommand{\Sref}[1]{{Sec.~\ref{#1}}}
\newcommand{\Eref}[1]{{Eq.~\ref{#1}}}
\newcommand{\Fref}[1]{{Fig.~\ref{#1}}}

\newcommand{\tf}[2]{\ensuremath{^{\text{#1}}\mat{T}^{}_{\text{#2}}}}
\newcommand{\tfinv}[2]{\ensuremath{^{\text{#1}}\mat{T}^{-1}_{\text{#2}}}}

\begin{document}

%TODO: title
%\title{A patient, a Surgeon, and a C-arm meet in the OR}
% subtitle:
\title{Closing the Calibration Loop: An Inside-out-tracking Paradigm for Augmented Reality in Orthopedic Surgery}
\titlerunning{Closing the Loop}  % abbreviated title (for running head)
%TODO: 

\author{Jonas~Hajek\inst{1,2,}\thanks{These authors have contributed equally.} \and Mathias~Unberath\inst{1,\star} \and Javad~Fotouhi\inst{1,\star} \and Bastian~Bier\inst{1,2} \and Sing~Chun~Lee\inst{1} \and Greg~Osgood\inst{3} \and Andreas~Maier\inst{2} \and Mehran~Armand\inst{4} \and Nassir~Navab\inst{1}}
\authorrunning{Hajek, Unberath, and Fotouhi et al.} % abbreviated author list (for running head)
 \institute{Computer Aided Medical Procedures, Johns Hopkins University\\
  \and
  Pattern Recognition Lab, Friedrich-Alexander-Universit{\"a}t Erlangen-N{\"u}rnberg
  \and
  Department of Orthopaedic Surgery, Johns Hopkins Hospital
  \and
  Applied Physics Laboratory, Johns Hopkins University
  }

\maketitle              % typeset the title of the contribution

\begin{abstract}

In percutaneous orthopedic interventions the surgeon attempts to reduce and fixate fractures in bony structures. The complexity of these interventions arises when the surgeon performs the challenging task of navigating surgical tools percutaneously only under the guidance of 2D interventional X-ray imaging. Moreover, the intra-operatively acquired data is only visualized indirectly on external displays. In this work, we propose a flexible Augmented Reality (AR) paradigm using optical see-through head mounted displays. The key technical contribution of this work includes the marker-less and dynamic tracking concept which closes the calibration loop between patient, C-arm and the surgeon. This calibration is enabled using Simultaneous Localization and Mapping of the environment of the operating theater. In return, the proposed solution provides 
\emph{in situ} visualization of pre- and intra-operative 3D medical data directly at the surgical site. We demonstrate pre-clinical evaluation of a prototype system, and report errors for calibration and target registration. Finally, we demonstrate the usefulness of the proposed inside-out tracking system in achieving ''bull's eye'' view for C-arm-guided punctures. This AR solution provides an intuitive visualization of the anatomy and can simplify the hand-eye coordination for the orthopedic surgeon.   

\end{abstract}
\section{Introduction}
Modern orthopedic trauma surgery focuses on percutaneous alternatives to many complicated procedures~\cite{gay1992,hong2010percutaneous}. These minimally invasive approaches are guided by intra-operative X-ray images that are acquired using mobile, non-robotic C-arm systems. It is well known that X-ray images from multiple orientations are required to warrant understanding of the 3D spatial relations since 2D fluoroscopy suffers from the effects of projective transformation. % try find better word: transmission and no occlusion
Mastering the mental mapping of tools to anatomy from 2D images is a key competence that surgeons acquire through extensive training. Yet, this task often challenges even experienced surgeons leading to longer procedure times, increased radiation dose, multiple tool insertions, and surgeon frustration~\cite{markelj2012review,andress2018onthefly}.\\
If 3D pre- or intra-operative imaging is available, challenges due to indirect visualization can be mitigated substantially reducing surgeon task load and fostering improved surgical outcome. Unfortunately, most of the previously proposed systems provide 3D information at the cost of integrating outside-in tracking solutions that require additional markers and intra-operative calibration that hinder clinical acceptance~\cite{markelj2012review}. As an alternative, intuitive and real-time visualization of 3D data in Augmented Reality (AR) environments has recently received considerable attention~\cite{andress2018onthefly,tuckertowards}. 
In this work, we present a purely image-based inside-out tracking concept and prototype system that dynamically closes the calibration loop between surgeon, patient, and C-arm enabling intra-operative optical see-through head-mounted display (OST HMD)-based AR visualization overlaid with the anatomy of interest. 
Such \emph{in situ} visualization could benefit residents in training that observe surgery to fully understand the actions of the lead surgeon with respect to the deep-seated anatomical targets. These applications in addition to simple task such as optimal positioning of C-arm systems, do not require the accuracy needed for surgical navigation and, therefore, could be the first target for OST HMD visualization in surgery. 
To the best of our knowledge, this prototype constitutes the first marker-less solution to intra-operative 3D AR on the target anatomy.

\section{Materials and Methods}
\begin {figure}[tb]
\centerline{\includegraphics[width=0.9\linewidth]{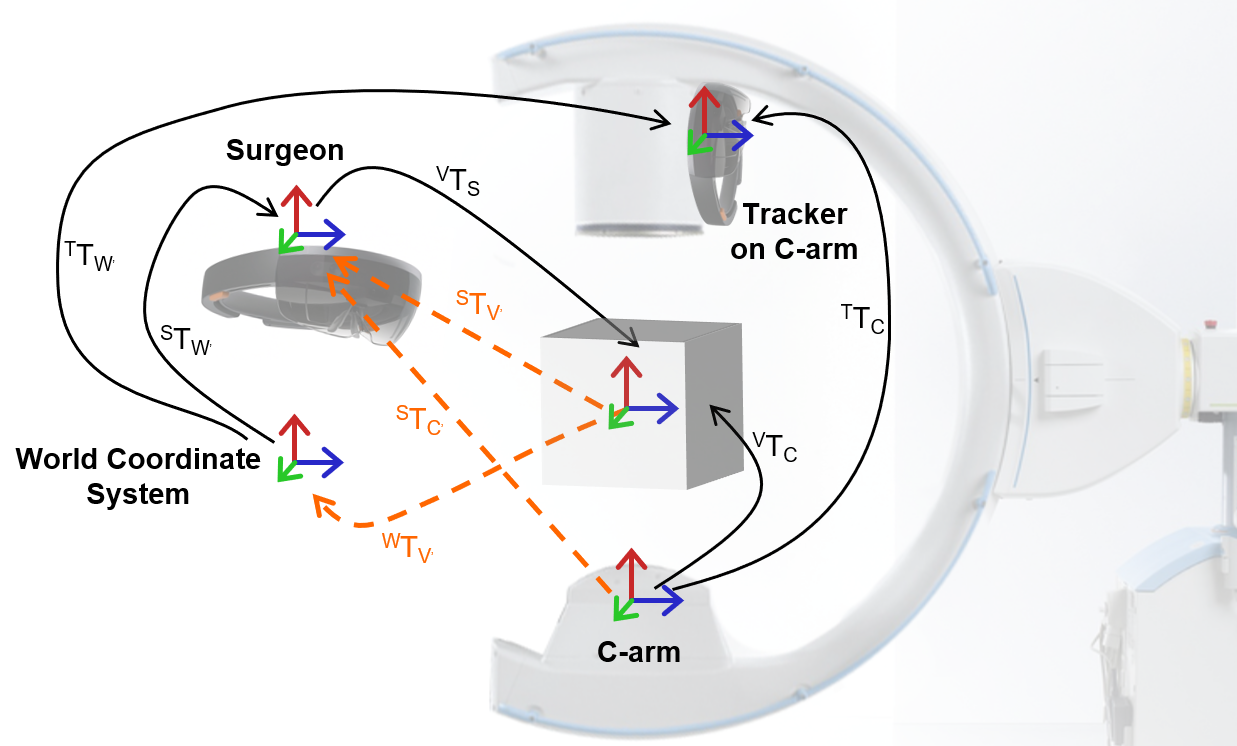}}
\caption{Spatial relations that are required to be estimated dynamically to enable the proposed AR environment. Transformations shown in black are estimated directly while transformations shown in orange are derived.}
\label{fig:transformations}
\end {figure}

\subsection{Calibration}
\label{sec:calibration}
The inside-out tracking paradigm, core of the proposed concept, is driven by the observation that all relevant entities (surgeon, patient, and C-arm) are positioned relative to the same environment, which we will refer to as the ''world coordinate system''. For intra-operative visualization of 3D volumes overlaid with the patient, we seek to dynamically recover 
\begin{equation}
\tf{S}{V}(t) = \tf{S}{W} \underbrace{\left( \tfinv{T}{W}(t_0)\ \tf{T}{C}(t_0) \right)\ \tfinv{V}{C}}_{\tf{W}{V}}\,,
\label{eq:surgeon2volume}
\end{equation}
the transformation describing the mapping from the surgeon's eyes to the 3D image volume. In \Eref{eq:surgeon2volume}, $t_0$ describes the time of pre- to intra-operative image registration while $t$ is the current time point. The spatial relations required to dynamically estimate $\tf{S}{V}$ are explained in the remainder of this section and visualized in \Fref{fig:transformations}.

\paragraph{$\tf{W}{S/T}$\,:} The transformations $\tf{W}{S/T}$ are estimated using Simultaneous Localization and Mapping (SLAM) thereby incrementally constructing a map of the environment, \ie the world coordinate system~\cite{endres2012evaluation}. Exemplarily for the surgeon, SLAM solves
\begin{equation}
\tf{W}{S}(t) = \underset{^{\text{W}}\mat{\hat{T}}_{\text{S}}}{\argmin}\ \mathrm{d}\!\left(  \mat{f}_{\text{W}}\left(\mat{P}\, ^{\text{W}}\mat{\hat{T}}_{\text{S}}(t)\mat{x}_{\text{S}}(t)\right), \mat{f}_{\text{S}}(t)   \right),
\end{equation}
where $\mat{f}_{\text{S}}(t)$ are features in the image at time $t$, $\mat{x}_{\text{S}}(t)$ are the 3D locations of these feature estimates either via depth sensors or stereo, $\mat{P}$ is the projection operator, and $d(\cdot,\cdot)$ is the feature similarity to be optimized. A key innovation of this work is the inside-out SLAM-based tracking of the C-arm w.\,r.\,t. the environment map by means of an additional tracker rigidly attached to the C-shaped gantry. This becomes possible if both trackers observe partially overlapping parts of the environment, \ie a feature rich and temporally stable area of the environment. This suggests, that the cameras on the C-arm tracker (in contrast to previous solutions~\cite{fotouhi2017pose,tuckertowards}) need to face the room rather than the patient.

\paragraph{$\tf{T}{C}$\,:} The tracker is rigidly mounted on the C-arm gantry suggesting that one-time offline calibration is possible. Since the X-ray and tracker cameras have no overlap, methods based on multi-modal patterns as in \cite{fotouhi2017pose,tuckertowards,andress2018onthefly} fail. However, if poses of both cameras w.\,r.\,t. the environment and the imaging volume, respectively, are known or can be estimated, Hand-Eye calibration is feasible~\cite{tsai1989new}. Put concisely, we estimate a rigid transform $\tf{T}{C}$ such that {$\mat{A}(t_i)\,\tf{T}{C} =\, \tf{T}{C}\mat{B}(t_i)$}, where $(\mat{A}/\mat{B})(t_i)$ is the relative pose between subsequent poses at times $i,i+1$ of the tracker and the C-arm, respectively. Poses of the C-arm $\tf{V}{C}(t_i)$ are known because our prototype (\Sref{sec:prototype}) uses a cone-beam CT (CBCT) enabled C-arm with pre-calibrated circular source trajectory such that several poses $\tf{V}{C}$ are known. During one sweep, we estimate the poses of the tracker $\tf{W}{T}(t_i)$ via \Eref{eq:surgeon2volume}. Finally, we recover $\tf{T}{C}$, and thus $\tf{W}{C}$, as detailed in~\cite{tsai1989new}.

\paragraph{$\tf{V}{C}$\,:} To close the loop by calibrating the patient to the environment, we need to estimate the $\tf{V}{C}$ describing the transformation from 3D image volumes to an intra-operatively acquired X-ray image. For pre-operative data, $\tf{V}{C}$ can be estimated via image-based 3D/2D registration, \eg as in~\cite{berger2016marker,de20163d}. If the C-arm is CBCT capable and the 3D volume is acquired intra-procedurally, $\tf{V}{C}$ is known and can be defined as one of the pre-calibrated C-arm poses on the source trajectory, \eg the first one. Once $\tf{V}{C}$ is known, the volumetric images are calibrated to the room via $\tf{W}{V}=\,\tf{W}{T}(t_0)\,\tf{T}{C}\,\tfinv{V}{C}(t_0)$, where $t_0$ denotes the time of calibration.

\subsection{Prototype}
\label{sec:prototype}
\begin {figure}
\centerline{\includegraphics[width=1.0\linewidth]{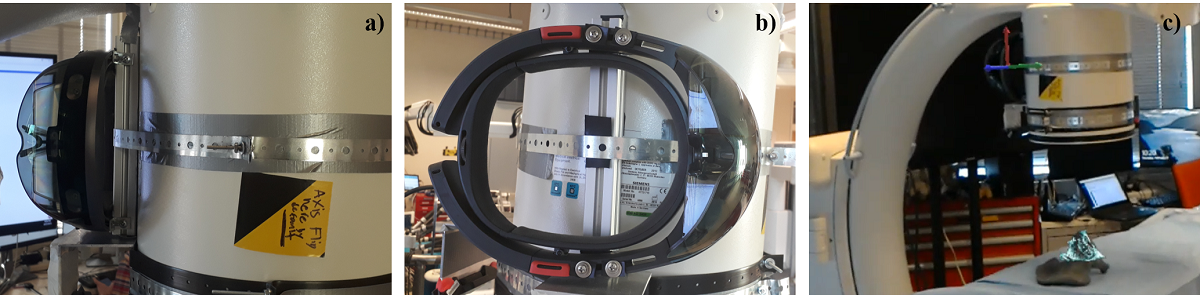}}
\caption{The prototype uses a Microsoft Hololens as tracker which is rigidly mounted on the C-arm detector, as demonstrated in Figs.~a) and b). In c), the coordinate axis of the RGB tracker is shown in relation to the mobile C-arm.}
\label{fig:prototype}
\end {figure}

For visualization of virtual content we use the Microsoft HoloLens (Microsoft, Redmond, WA) that simultaneously serves as inside-out tracker providing $\tf{W}{S}$ according to see \Sref{sec:calibration}. To simplify communication between devices, we mount a second HoloLens device on C-arm to track movement of the gantry $\tf{W}{T}$. We use a CBCT enabled mobile C-arm (Siemens Arcadis Orbic 3D, Siemens Healthineers, Forchheim, Germany) and rigidly attach the tracking device to the image intensifier with the principal ray of the front facing RGB camera oriented parallel to the patient table as demonstrated in \Fref{fig:prototype}.
$\tf{T}{C}$ is estimated via Hand-Eye calibration from 98 (tracker, C-arm) absolute pose pairs acquired during a circular source trajectory yielding 4753 relative poses. Since the C-arm is CBCT enabled, we simplify estimation of $\tf{V}{C}$ and define $t_0$ to correspond to the first C-arm pose.

\subsection{Virtual Content in the AR Environment}
\begin {figure}[tb]
\centerline{\includegraphics[width=1\linewidth]{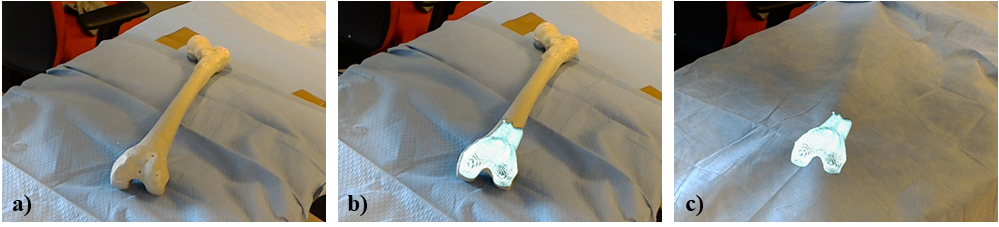}}
\caption{Since $\tf{S}{V}(t)$ is known, the real object in a) is overlaid with the rendered volume shown in b). Correct overlay persists even if the real object is covered in c). }
\label{fig:bonerendering}
\end {figure}

\begin {figure}[tb]
\centerline{\includegraphics[width=1\linewidth]{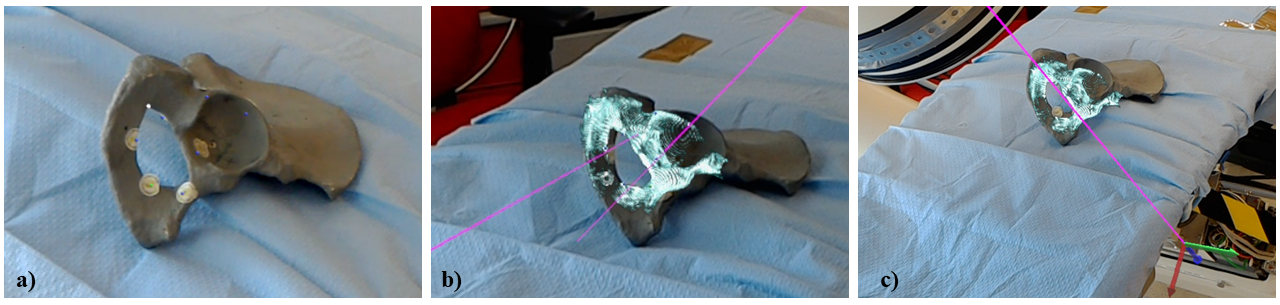}}
\caption{a) Pelvis phantom used for TRE assessment. b) Lines placed during the experiment to evaluate point-to-line TRE. c) Visualization of the X-ray source and principal ray next to the same phantom.}
\label{fig:arenvironment}
\end {figure}

Once all spatial relations are estimated, multiple augmentations of the scene become possible. We support visualization of the following content depending on the task (see \Fref{fig:arenvironment}): Using $\tf{S}{V}(t)$ we provide volume renderings of the 3D image volumes overlaid with the patient's anatomy as shown in \Fref{fig:bonerendering}. In addition to the volume rendering, annotations of the 3D data (such as landmarks) can be displayed. Further and via $\tf{S}{C}(t)$, the C-arm source and principal ray, seen in \Fref{fig:arenvironment} c) can be visualized as the C-arm gantry is moved to different viewing angles. Volume rendering and principal ray visualization combined are an effective solution to determine ''bull's eye'' views to guide punctures~\cite{morimoto2010c}. 

\subsection{Experiments and Feasibility Study}

\paragraph{Hand-Eye Residual Error:}
Following \cite{tsai1989new}, we compute the rotational and translational component of $\tf{T}{C}$ independently. Therefore, we state the residual of solving $\mat{A}(t_i)\,\tf{T}{C} =\, \tf{T}{C}\mat{B}(t_i)$ for $\tf{T}{C}$ separately for rotation and translation averaged over all relative poses.

\paragraph{Target Registration Error:} We evaluate the end-to-end target registration error (TRE) of our prototype system using a Sawbones phantom (Sawbones, Vashon, WA) with metal spheres on the surface. The spheres are annotated in a CBCT of the phantom and serve as the targets for TRE computation. 
Next, $M=4$ medical experts are asked to locate the spheres in the AR environment: For every of the $N=7$ spheres $\mat{p}_i$, the user $j$ changes position in the room, and using the ''air tap'' gesture defines a 3D line $\mat{l}^j_i$ corresponding to his gaze that intersects the sphere on the phantom. The TRE is then defined as 
\begin{equation}
\text{TRE} = \frac{1}{M\cdot N }\sum_{j=1}^{M}\sum_{i=1}^{N} \mathrm{d}(\mat{p}_i,\mat{l}^j_i)\,,
\end{equation}
where $\mathrm{d}(\mat{p},\mat{l})$ is the 3D point-to-line distance.

\paragraph{Achieving ''Bull's Eye'' View:} Complementary to the technical assessment, we conduct a clinical task-based evaluation of the prototype: Achieving ''bull's eye'' view for percutaneous punctures. To this end, we manufacture cubic foam phantoms and embed a radiopaque tubular structure (radius $\approx5$\,mm) at arbitrary orientation but invisible from the outside. A CBCT is acquired and rendered in the AR environment overlaid with the physical phantom such that the tube is clearly discernible. Further, the principal ray of the C-arm system is visualized. Again, $M=4$ medical experts are asked to move the gantry such that the principal ray pierces the tubular structure, thereby achieving the desired ''bull's eye'' view. Verification of the view is performed by acquiring an X-ray image. Additionally, users advance a K-wire through the tubular structure under ''bull's eye'' view guidance using X-rays from the view selected in the AR environment. Placement of the K-wire without breaching of the tube is verified in the guidance and a lateral X-ray view.

\section{Results}

\paragraph{Hand-Eye Residual Error:} We quantified the residual error of our Hand-Eye calibration between the C-arm and tracker separately for rotational and translational component. For rotation, we found an average residual of $6.18^\circ$, $5.82^\circ$, and $5.17^\circ$ around $\vec{e}_x$, $\vec{e}_y$, and $\vec{e}_z$, respectively, while for translation the root-mean-squared residual was $26.6$\,mm. It is worth mentioning that the median translational error in $\vec{e}_x$, $\vec{e}_y$, and $\vec{e}_z$ direction was $4.10$\,mm, $3.02$\,mm, $43.18$\,mm, respectively, where $\vec{e}_z$ corresponds to the direction of the principal ray of the tracker coordinate system, \ie the rotation axis of the C-arm.

\paragraph{Target Registration Error:} The point-to-line TRE averaged over all points and users was $11.46$\,mm.

\begin {figure}[tb]
\centerline{\includegraphics[width=1\linewidth]{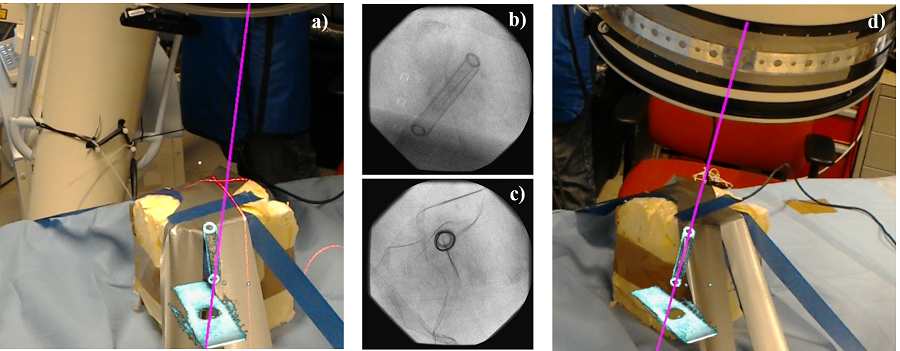}}
\caption{Screen captures from the user's perspective attempting to achieve the ''bull's eye'' view. The virtual line (purple) corresponds to the principal ray of the C-arm system in the current pose while the CBCT of the phantom is volume rendered in light blue. a) The C-arm is positioned in neutral pose with the laser cross-hair indicating that the phantom is within the field of view. The AR environment indicates mis-alignment for ''bull's eye'' view that is confirmed using an X-ray (b). After alignment of the virtual principal ray with the virtual tubular structure inside the phantom (d), an acceptable ''bull's eye'' view is achieved (c).}
\label{fig:bullseye}
\end {figure}

\paragraph{Achieving ''Bull's Eye'' View:} Every user successfully achieved a ''bull's eye'' view in the first try that allowed them to place a K-wire without breach of the tubular structure. \Fref{fig:bullseye} shows representative scene captures acquired from the perspective of the user. We would like to refer to the supplementary material where we provide a video documenting one trial from both a bystander's and the user's perspective.

% X-ray of Bull's eye? 

\section{Discussion and Conclusion}
We presented an inside-out tracking paradigm to close the transformation loop for AR in orthopedic surgery based upon the realization that surgeon, patient, and C-arm can be calibrated to the environment. Our entirely marker-less approach enables rendering of virtual content at meaningful positions, \ie dynamically overlaid with the patient and the C-arm source. The performance of our prototype system is promising and enables effective ''bull's eye'' viewpoint planning for punctures.\\
Despite an overall positive evaluation, some limitations remain. The TRE of $11.46$\,mm is acceptable for viewpoint planning, but may be unacceptably high if the aim of augmentation is direct feedback on tool trajectories as in~\cite{andress2018onthefly,tuckertowards}. The TRE is compound of multiple sources of error: 1) Residual errors in Hand-Eye calibration of $\tf{T}{C}$, particularly due to the fact that poses are acquired on a circular trajectory and are, thus, co-planar as supported by our quantitative results; and 2) Inaccurate estimates of $\tf{W}{T}$ and $\tf{W}{S}$ that indirectly affect all other transformations. We anticipate improvements in this regard when additional out-of-plane pose pairs are sampled for Hand-Eye calibration.
Further, the accuracy of estimating $\tf{W}{T/S}$ is currently limited by the capabilities of Microsoft's HoloLens and is expected to improve in the future.  \\  
%In our prototype, we valued the convenience of calibration-free tracking w.\,r.\,t. the same environment map higher than smaller tracking errors achievable with more dedicated systems. For future prototypes directed towards clinical evaluation, use of more powerful trackers at the cost of additional calibration may become necessary.\\
In summary, we believe that our approach has great potential to benefit orthopedic trauma procedures particularly when pre-operative 3D imaging is available.
%We expect our marker-less approach to only marginally affect the clinical work flow and, therefore, reach a higher clinical acceptance than current marker based solutions. 
In addition to the benefits for the surgeon discussed here, the proposed AR environment may prove beneficial in an educational context where residents must comprehend the lead surgeon's actions. 
%In this regard our marker-less approach stands out by maintaining the clinical work flow.
Further, we envision scenarios where the proposed solution can support the X-ray technician in achieving the desired views of the target anatomy.

%\bibliographystyle{splncs}
%\bibliography{refs}

\begin{thebibliography}{10}
	
	\bibitem{gay1992}
	Gay, B., Goitz, H.T., Kahler, A.:
	\newblock {Percutaneous CT Guidance : Screw Fixation of Acetabular Fractures
		Preliminary Results of a New Technique with}.
	\newblock American Journal of roentgenology \textbf{158}(4) (1992)  819--822
	
	\bibitem{hong2010percutaneous}
	Hong, G., Cong-Feng, L., Cheng-Fang, H., Chang-Qing, Z., Bing-Fang, Z.:
	\newblock Percutaneous screw fixation of acetabular fractures with 2d
	fluoroscopy-based computerized navigation.
	\newblock Archives of orthopaedic and trauma surgery \textbf{130}(9) (2010)
	1177--1183
	
	\bibitem{markelj2012review}
	Markelj, P., Toma{\v{z}}evi{\v{c}}, D., Likar, B., Pernu{\v{s}}, F.:
	\newblock {A review of 3D/2D registration methods for image-guided
		interventions}.
	\newblock Med Image Anal \textbf{16}(3) (2012)  642--661
	
	\bibitem{andress2018onthefly}
	Andress, S., Johnson, A., Unberath, M., Winkler, A., Yu, K., Fotouhi, J.,
	Weidert, S., Osgood, G., Navab, N.:
	\newblock On-the-fly augmented reality for orthopedic surgery using a
	multimodal fiducial.
	\newblock Journal of Medical Imaging \textbf{5} (2018)  5 -- 5 -- 12
	
	\bibitem{tuckertowards}
	Tucker, E., Fotouhi, J., Lee, S., Unberath, M., Fuerst, B., Johnson, A.,
	Armand, M., Osgood, G., Navab, N.:
	\newblock {Towards clinical translation of augmented orthopedic surgery: from
		pre-op CT to intra-op X-ray via RGBD sensing}.
	\newblock In: SPIE Medical Imaging. (2018)
	
	\bibitem{endres2012evaluation}
	Endres, F., Hess, J., Engelhard, N., Sturm, J., Cremers, D., Burgard, W.:
	\newblock An evaluation of the rgb-d slam system.
	\newblock In: Robotics and Automation (ICRA), 2012 IEEE International
	Conference on, IEEE (2012)  1691--1696
	
	\bibitem{fotouhi2017pose}
	Fotouhi, J., Fuerst, B., Johnson, A., Lee, S.C., Taylor, R., Osgood, G., Navab,
	N., Armand, M.:
	\newblock {Pose-aware C-arm for automatic re-initialization of interventional
		2D/3D image registration}.
	\newblock International Journal of Computer Assisted Radiology and Surgery
	\textbf{12}(7) (2017)  1221--1230
	
	\bibitem{tsai1989new}
	Tsai, R.Y., Lenz, R.K.:
	\newblock {A new technique for fully autonomous and efficient 3D robotics
		hand/eye calibration}.
	\newblock IEEE Transactions on Robotics and Automation \textbf{5}(3) (1989)
	345--358
	
	\bibitem{berger2016marker}
	Berger, M., M{\"u}ller, K., Aichert, A., Unberath, M., Thies, J., Choi, J.H.,
	Fahrig, R., Maier, A.:
	\newblock {Marker-free motion correction in weight-bearing cone-beam CT of the
		knee joint}.
	\newblock Medical Physics \textbf{43}(3) (2016)  1235--1248
	
	\bibitem{de20163d}
	De~Silva, T., Uneri, A., Ketcha, M., Reaungamornrat, S., Kleinszig, G., Vogt,
	S., Aygun, N., Lo, S., Wolinsky, J., Siewerdsen, J.:
	\newblock {3D--2D image registration for target localization in spine surgery:
		investigation of similarity metrics providing robustness to content
		mismatch}.
	\newblock Physics in Medicine \& Biology \textbf{61}(8) (2016)  3009
	
	\bibitem{morimoto2010c}
	Morimoto, M., Numata, K., Kondo, M., Nozaki, A., Hamaguchi, S., Takebayashi,
	S., Tanaka, K.:
	\newblock {C-arm cone beam CT for hepatic tumor ablation under real-time 3D
		imaging}.
	\newblock American Journal of Roentgenology \textbf{194}(5) (2010)  W452--W454
	
\end{thebibliography}

\end{document}